%% file: main.tex
\documentclass[10pt,twocolumn,letterpaper]{article}

\usepackage[pagenumbers]{cvpr}

\definecolor{cvprblue}{rgb}{0.21,0.49,0.74}
\usepackage[pagebackref,breaklinks,colorlinks,allcolors=cvprblue]{hyperref}

\def\paperID{*****}
\def\confName{CVPR}
\def\confYear{2026}

\title{Context as Prior: Bayesian-Inspired Intent Inference for Non-Speaking Agents with a Household Cat Testbed}

\author{Wenqian Zhang\\
University of California, Riverside\\
Riverside, CA\\
{\tt\small wzhan240@ucr.edu}
\and
Zehao Wang\\
University of California, Riverside\\
Riverside, CA\\
{\tt\small zwang712@ucr.edu}
}

\begin{document}
\maketitle

\begin{abstract}
Many agents in real-world environments cannot reliably communicate their goals through language, including household pets, pre-verbal infants, and other non-speaking embodied agents. In such settings, intent must be inferred from incomplete behavioral observations in context-rich environments. This creates a core ambiguity: observable behavior is often noisy or underspecified, while context provides strong prior information but can also induce brittle shortcut predictions if used naively.

We present \textbf{CatSignal}, a Bayesian-inspired probabilistic framework for multimodal intent inference that models spatial context as a prior-like constraint and behavioral observations as evidence. Rather than treating context as an ordinary input feature, our method uses a context-gated Product-of-Experts formulation to compute posterior-like intent distributions from context, pose dynamics, and acoustic cues. We instantiate this formulation in a household cat setting as a focused proof-of-concept for intent inference in non-speaking agents.

Under Leave-One-Video-Out evaluation on a multimodal domestic cat dataset, the proposed prior-guided fusion achieves the best overall accuracy of \textbf{77.72\%}, outperforming feature concatenation (71.83\%) and stronger late-fusion baselines. More importantly, it substantially reduces context-driven shortcut failures in ambiguous cases. While simpler fusion strategies remain competitive in Macro-F1 and selective prediction, the proposed model provides the strongest overall accuracy and the best suppression of context-based shortcut collapse.
\end{abstract}

% ------------------------------------------------
\section{Introduction}

Intelligent systems operating in homes and shared environments increasingly need to interpret intents that cannot reliably communicate through natural language, such as household pets, pre-verbal infants, and other non-speaking companions~\cite{premack1978,baker2009,baker2011bayesian,rabinowitz2018machine,vanhorenbeke2021activity}. In such settings, intent must be inferred from partial observations rather than explicit verbal instruction. This is challenging as behavioral cues are often noisy or incomplete, while environmental context is highly informative and often crucial for narrowing the set of feasible goals.

A domestic cat provides a concrete example. A cat near a food bowl may be seeking food, but it may also simply remain idle nearby. Likewise, a cat near a door may intend to exit, or may merely linger there. Pose and audio alone are often insufficient to resolve such ambiguity: motion patterns can overlap across goals, and vocalizations are sparse and inconsistent~\cite{berman2014mapping,wiltschko2015mapping,bohnslav2021deepethogram}. Meanwhile, context provides a strong prior over feasible goals, but cannot by itself determine whether the observed clip reflects active goal pursuit or context-compatible idling.

If context is treated as an ordinary feature, a discriminative model may collapse to shortcut rules such as \textit{near bowl} $\rightarrow$ \textit{FOOD} or \textit{near door} $\rightarrow$ \textit{EXIT}, regardless of behavioral evidence~\cite{geirhos2020shortcut}. We therefore frame intent inference for non-speaking agents as probabilistic reasoning under strong contextual priors. We propose \textbf{CatSignal}, a Bayesian-inspired context-gated Product-of-Experts framework in which spatial context defines a prior-like constraint over feasible intents, while pose and audio provide complementary evidence~\cite{baltrusaitis2019multimodal,ngiam2011multimodal,hinton2002training,wu2018multimodal}.

We instantiate this formulation in a household feline setting as a proof-of-concept testbed~\cite{mathis2018deeplabcut,pereira2022sleap,chen2023mammalnet,wiltshire2023deepwild,vidal2021perspectives,ye2024superanimal}. Our contributions are: (1) a prior-guided formulation for intent inference in non-speaking agents; (2) a context-gated Product-of-Experts framework combining context, pose, and audio; and (3) experiments showing improved overall accuracy, reduced shortcut failures, and competitive uncertainty behavior under ambiguity.

\begin{figure*}[t] 
    \centering
    \includegraphics[width=0.85\textwidth]{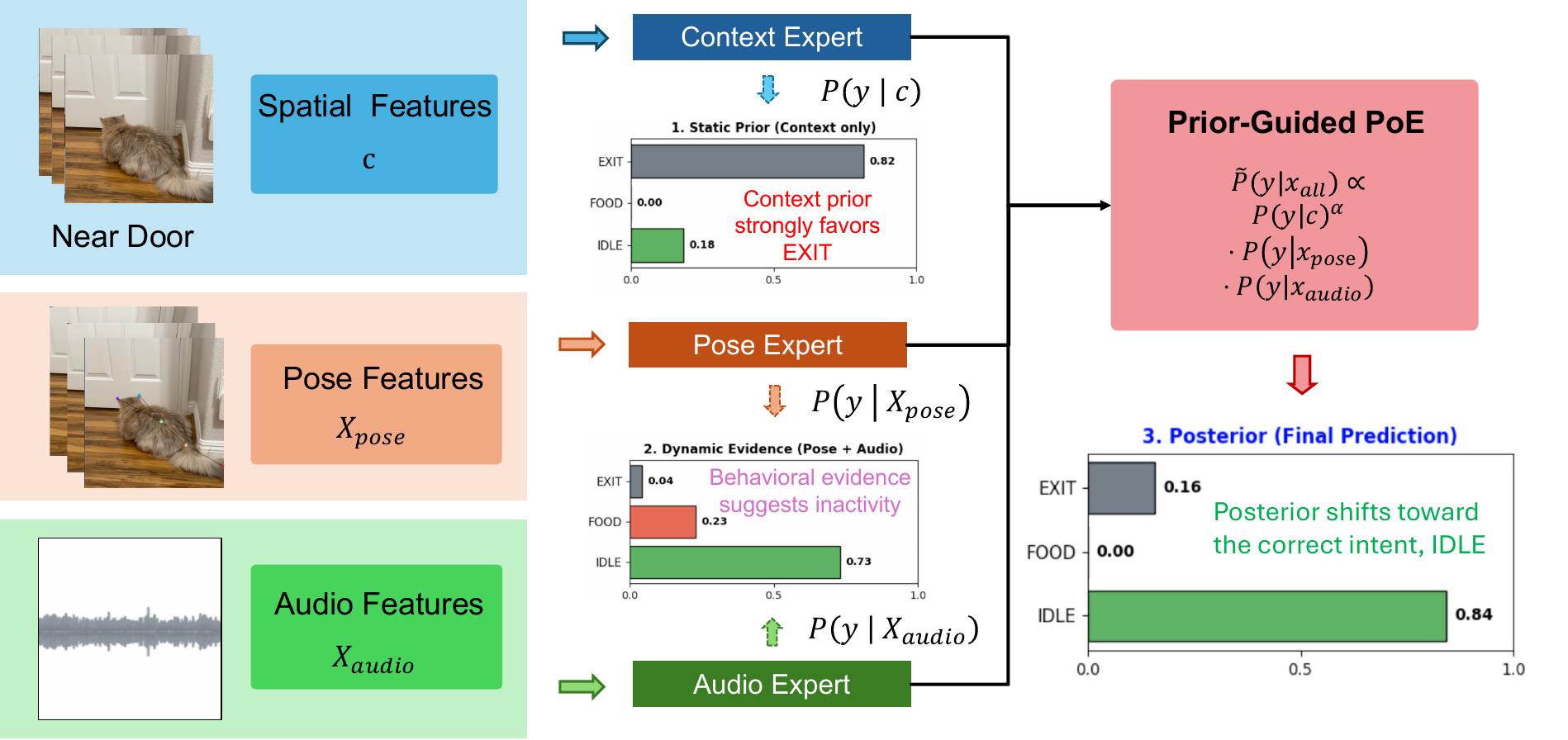}
    
    % \vspace{-0.3cm} 
    \caption{Illustration of prior-guided intent inference in an ambiguous near-door household-cat clip. Context induces a strong prior toward \textit{EXIT}, while pose and audio evidence favor \textit{IDLE}. Their Product-of-Experts fusion shifts the final posterior toward \textit{IDLE}.}
    \label{fig:pipeline}
    
    \vspace{-0.5cm} 
\end{figure*}

% ------------------------------------------------
\section{Related Work}

Markerless pose systems such as DeepLabCut~\cite{mathis2018deeplabcut} and SLEAP~\cite{pereira2022sleap} have enabled animal behavior analysis in naturalistic settings, while benchmarks such as MammalNet~\cite{chen2023mammalnet} reflect a growing interest in animal-centered visual understanding. Additional work on animal behavior quantification and ecological deployment further highlights the growth of this area~\cite{berman2014mapping,wiltschko2015mapping,bohnslav2021deepethogram,wiltshire2023deepwild,vidal2021perspectives,ye2024superanimal}. However, most prior work focuses on observable motion or action categories rather than latent goal inference.

More broadly, hidden-goal inference has been studied in Bayesian theory of mind~\cite{baker2011bayesian}, inverse planning~\cite{baker2009}, machine theory of mind~\cite{rabinowitz2018machine}, and goal/activity recognition more broadly~\cite{meneguzzi2021survey,vanhorenbeke2021activity}. In multimodal learning, feature concatenation is a common fusion strategy~\cite{baltrusaitis2019multimodal,ngiam2011multimodal}, but it can encourage shortcut reliance when the context is strongly predictive~\cite{geirhos2020shortcut}. Product-of-Experts formulations provide a natural way to combine probabilistic contributions from multiple modalities~\cite{hinton2002training,wu2018multimodal}. Our contribution is to treat context not as an ordinary feature, but as prior-like structure that is updated by behavioral evidence.

% ------------------------------------------------
\section{Dataset and Problem Setup}

We instantiate our formulation in a household feline setting as a focused proof-of-concept for intent inference in non-speaking agents. The raw dataset consists of 34 home videos of varying durations recorded in a real domestic environment, with a total duration of 1855.76 seconds. After pose-validity filtering, the final dataset contains 212 retained 3-second clips.

To construct training and evaluation samples, we segment the recordings into 3-second clips, but retain only clips with valid pose observations. Specifically, we first run DeepLabCut~\cite{mathis2018deeplabcut} and keep only temporal segments in which the cat is successfully detected with sufficient valid keypoints for extracting pose-derived features such as body speed, tail speed, tail swing, and body stretch statistics. We then extract synchronized audio from the same retained temporal windows, so that pose and audio evidence are aligned at the clip level. The final feature set, therefore, combines pose-validity statistics, motion descriptors, and synchronized acoustic descriptors for each retained clip.

\textit{Intent} refers to an inferred latent goal associated with the observed behavior, with labels assigned retrospectively based on the immediate behavioral outcome. We consider three intent classes: \textit{EXIT}, where the cat exhibits exit-seeking behavior and then leaves once the door is opened; \textit{FOOD}, where the cat exhibits food-seeking behavior and then consumes food; and \textit{IDLE}, where the cat remains inactive or does not realize an exit- or food-directed outcome.

The spatial context is represented by three discrete states: \texttt{near\_bowl}, \texttt{near\_door}, and \texttt{neutral}.
Context provides a strong spatial prior over feasible intents in this household environment. Exit-related behavior occurs only near the door, while food-seeking behavior occurs only near the bowl. In the retained dataset, the \texttt{near\_bowl} region contains only \textit{FOOD} and \textit{IDLE} clips, the \texttt{near\_door} region contains only \textit{EXIT} and \textit{IDLE} clips, and the \texttt{neutral} region contains only \textit{IDLE}. This makes the task fundamentally asymmetric: context strongly constrains which intents are feasible, but does not fully determine whether the observed clip reflects a goal-directed state or a context-compatible idle state. The core challenge is therefore not identifying a relevant context, but disambiguating goal-directed behavior from idling within that context.

% We evaluate all methods under Leave-One-Video-Out (LOVO) cross-validation to test generalization across recording sessions rather than within-video temporal interpolation.

% ------------------------------------------------
\section{Method}

Figure~\ref{fig:pipeline} illustrates our prior-guided intent inference framework through an ambiguous near-door example. In this clip, spatial context induces a strong prior toward EXIT, while pose and audio provide behavioral evidence favoring IDLE. More generally, given a video segment, our model extracts three streams of information: spatial context, pose dynamics, and acoustic features, each of which contributes a modality-specific class distribution. The final prediction is obtained through context-gated Product-of-Experts fusion.

\textbf{Context as prior.}
Let \( y \in \{\textit{EXIT}, \textit{FOOD}, \textit{IDLE}\} \) denote the latent intent. We interpret spatial context \( c \) as a prior-like constraint \(P(y \mid c)\). This prior captures coarse environmental feasibility constraints. For example, \(P(\textit{EXIT}\mid \texttt{near\_door})\) is expected to be high, while \(P(\textit{FOOD}\mid \texttt{near\_door})\) is negligible. However, the prior is not sufficient to determine whether the observed clip reflects active goal pursuit or an idle state within the same region.

\textbf{Behavioral evidence.}
Using DeepLabCut, we estimate 2D keypoints and extract kinematic descriptors such as body speed, body stretch, tail speed, and tail movement statistics, which define a pose expert \(P(y \mid x_{\text{pose}})\). From synchronized audio, we extract MFCC and spectral statistics to define an audio expert \(P(y \mid x_{\text{audio}})\).

\textbf{Prior-guided Product-of-Experts fusion.}
We combine prior and evidence using
\setlength{\abovedisplayskip}{3pt}
\setlength{\belowdisplayskip}{3pt}
\begin{equation}
    \tilde{P}(y \mid x_{\text{all}}) \propto P(y \mid c)^\alpha \cdot P(y \mid x_{\text{pose}})\cdot P(y \mid x_{\text{audio}}).
\end{equation}
where \(\alpha\) controls the strength of the contextual prior. Ablation results show that removing context severely harms performance, while moderate-to-full prior strength performs best; we therefore use \(\alpha=1.0\) in the main experiments. After normalization, this yields the posterior-like intent distribution used for prediction.

We tune the pose and audio experts while keeping the low-dimensional context expert fixed. Although the formulation is not a fully generative Bayesian model, it preserves the key distinction that context should act as prior structure rather than as an ordinary feature.
% ------------------------------------------------
\section{Experiments}

Our experiments test a central hypothesis: in a household setting with strong spatial priors, treating context as prior knowledge should yield more reliable intent inference than treating it as an ordinary feature.
We evaluate all methods under Leave-One-Video-Out (LOVO) cross-validation to test generalization across recording sessions rather than within-video temporal interpolation.

\subsection{Baselines}

We compare ours against \textbf{Pose-only}, \textbf{Audio-only}, \textbf{Context-only}, \textbf{Feature Concat}, \textbf{LateFusion-Avg}, \textbf{LateFusion-Weighted}, \textbf{PoE-Ctx+Pose}, and \textbf{PoE-Ctx+Aud}. These baselines range from single-modality prediction to stronger feature-, late-fusion, and partial-expert alternatives, and test whether our gains come from multimodal information alone or from assigning context a distinct prior-like role.

\subsection{Main Results}

Table~\ref{tab:main_results} reports performance. The full prior-guided PoE model achieves the best overall accuracy, improving from 71.83\% for feature concatenation to \textbf{77.72\%}. It also outperforms both late-fusion and partial-expert baselines in accuracy, indicating that treating context as a distinct prior-like factor can be beneficial when combined with multimodal evidence.
% \begin{table}[t]
%     \centering
%     \begin{tabular}{lcc}
%         \toprule
%         Method & Accuracy (\%) & Macro-F1 \\
%         \midrule
%         Context-only & 60.87 & 0.7117 \\
%         Feature Concat & 71.83 & 0.6666 \\
%         LateFusion-Avg & 73.69 & \textbf{0.7666}\\
%         PoE-Ctx+Aud & 75.31 & 0.7549 \\
%         \textbf{Prior-Guided PoE (Full)} & \textbf{77.72} & 0.7460 \\
%         \bottomrule
%     \end{tabular}
%     \caption{Overall intent prediction performance under LOVO evaluation. The full prior-guided PoE model achieves the highest overall accuracy.}
%     \label{tab:main_results}
% \end{table}
\begin{table}[t]
    \centering
    \footnotesize
    \begin{tabular}{lccc}
        \toprule
        Method & Acc. (\%) & Std. (\%) & Macro-F1 \\
        \midrule
        Context-only & 60.87 & 48.80 & 0.7117 \\
        Audio-only & 50.87 & 38.76 & 0.4977 \\
        Pose-only & 39.36 & 36.25 & 0.3623 \\
        \midrule
        Feature Concat & 71.83 & 35.35 & 0.6666 \\
        LateFusion-Avg & 73.69 & 33.88 & \textbf{0.7666} \\
        LateFusion-Weighted & 73.76 & 32.02 & 0.7454 \\
        \midrule
        PoE-Ctx+Pose & 60.55 & 43.56 & 0.6729 \\
        PoE-Ctx+Aud & 75.31 & 35.77 & 0.7549 \\
        \textbf{Prior-Guided PoE (Full)} & \textbf{77.72} & 33.24 & 0.7460 \\
        \bottomrule
    \end{tabular}
    \caption{Overall intent prediction performance under LOVO evaluation. The full prior-guided PoE model achieves the highest overall accuracy.}
    \label{tab:main_results}
\end{table}
These results show that neither pose nor audio is sufficient in isolation, while context is strong but incomplete because it cannot distinguish active goal from context-compatible idling. The best performance is obtained when context is treated as a prior and updated by behavioral evidence rather than fused naively as an ordinary feature. 

\subsection{Ablation on Prior Strength}

Table~\ref{tab:alpha_ablation} shows the effect of varying the context-strength parameter \(\alpha\). Removing context severely harms performance, while moderate-to-full prior strength performs best. In particular, \(\alpha=1.0\) yields the highest overall accuracy, which we therefore use in the main experiments.

% \begin{table}[t]
%     \centering
%     \begin{tabular}{cccc}
%         \toprule
%         \(\alpha\) & Acc. (\%) & Std. (\%) & Macro-F1 \\
%         \midrule
%         0.0 & 45.60 & \textbf{37.01} & 0.4156 \\
%         0.5 & 75.42 & 34.71 & 0.6922 \\
%         0.8 & 75.75 & 34.14 & 0.7100 \\
%         \textbf{1.0} & \textbf{77.72} & 33.24 & \textbf{0.7460} \\
%         1.2 & 75.11 & 34.27 & 0.7244 \\
%         \bottomrule
%     \end{tabular}
%     \caption{Sensitivity of the full PoE model to the context-strength parameter \(\alpha\) under tuned experts.}
%     \label{tab:alpha_ablation}
% \end{table}

\begin{table}[t]
    \centering
    \footnotesize
    \begin{tabular}{lccc}
        \toprule
        $\alpha$ & Acc. (\%) & Std. (\%) & Macro-F1 \\
        \midrule
        0.0 & 45.60 & 37.01 & 0.4156 \\
        0.3 & 73.10 & 35.19 & 0.6577 \\
        0.5 & 75.42 & 34.71 & 0.6922 \\
        0.8 & 75.75 & 34.14 & 0.7100 \\
        \textbf{1.0} & \textbf{77.72} & 33.24 & \textbf{0.7460} \\
        1.2 & 75.11 & 34.27 & 0.7244 \\
        \bottomrule
    \end{tabular}
    \caption{Sensitivity of the full PoE model to the context-strength parameter $\alpha$ under tuned experts.}
    \label{tab:alpha_ablation}
\end{table}

% ------------------------------------------------
\section{Analysis}

Because the main motivation of our method is not merely higher average accuracy, but more reliable inference under ambiguity, we focus on analyses that directly probe shortcut behavior and confidence quality.

\subsection{Shortcut Failures under Ambiguous Contexts}

\begin{figure}[t]
    \centering
    \includegraphics[width=0.95\linewidth]{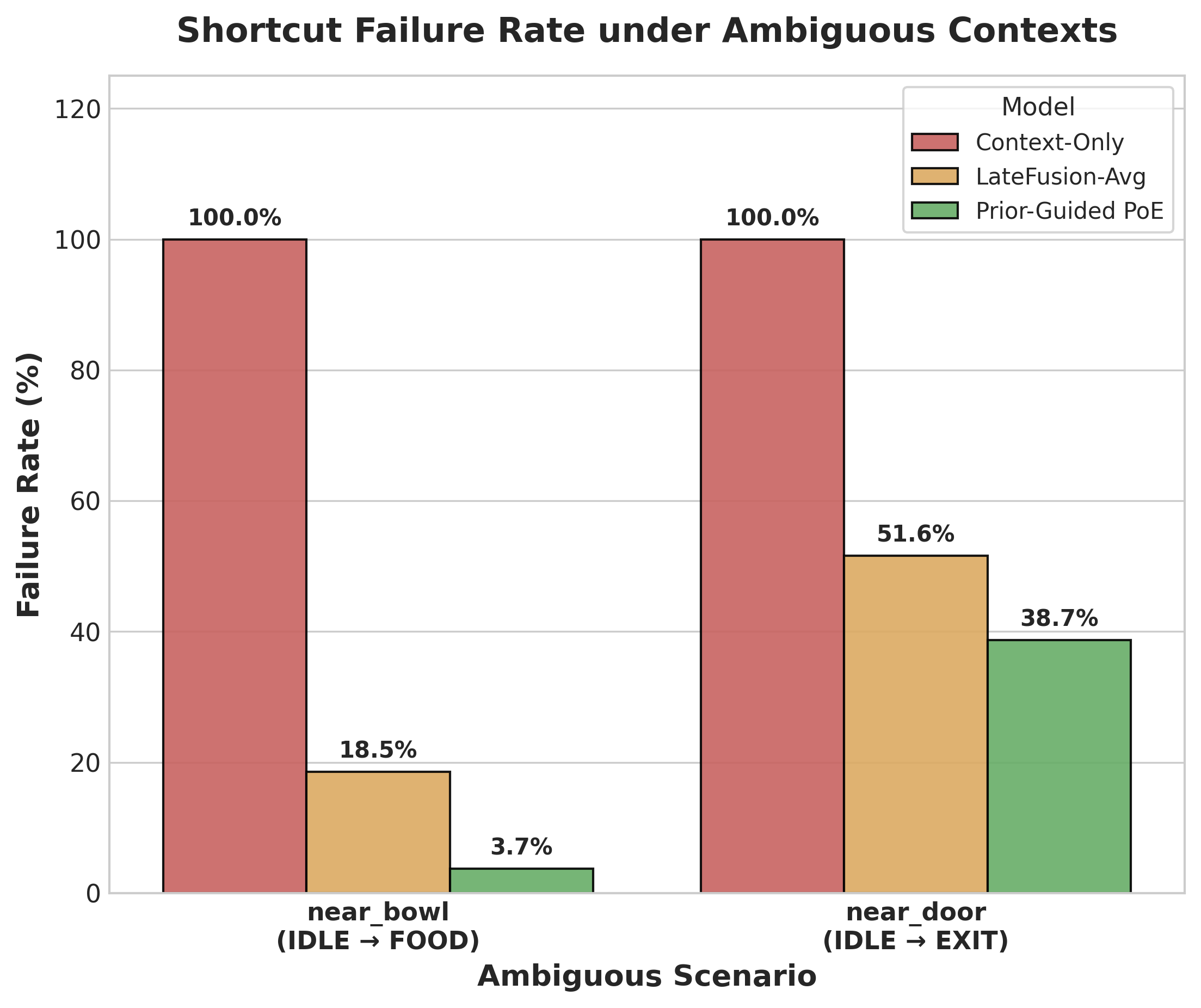}
    \caption{Shortcut failure rates in ambiguous contexts under tuned experts. Context-only collapses to deterministic shortcuts, while late fusion and prior-guided PoE reduce such failures.}
    \label{fig:shortcut_failure}
\end{figure}

Figure~\ref{fig:shortcut_failure} highlights the central failure mode in this task: deterministic shortcut predictions induced by context alone. In ambiguous scenarios, the context-only model collapses to context-based shortcuts, with 100\% failure on idle samples in both ambiguous settings. Both late fusion and prior-guided PoE substantially reduce these failures. Under the tuned experts, the proposed PoE is especially effective in the near-bowl regime, reducing the shortcut failure rate for \textit{IDLE} $\rightarrow$ \textit{FOOD} errors from 18.5\% under late fusion to 3.7\%. In the near-door regime, PoE remains better than late fusion, though the margin is smaller (38.7\% vs.\ 51.6\%).

\subsection{Selective Prediction under Ambiguity}

\begin{figure}[t]
    \centering
    \includegraphics[width=\linewidth]{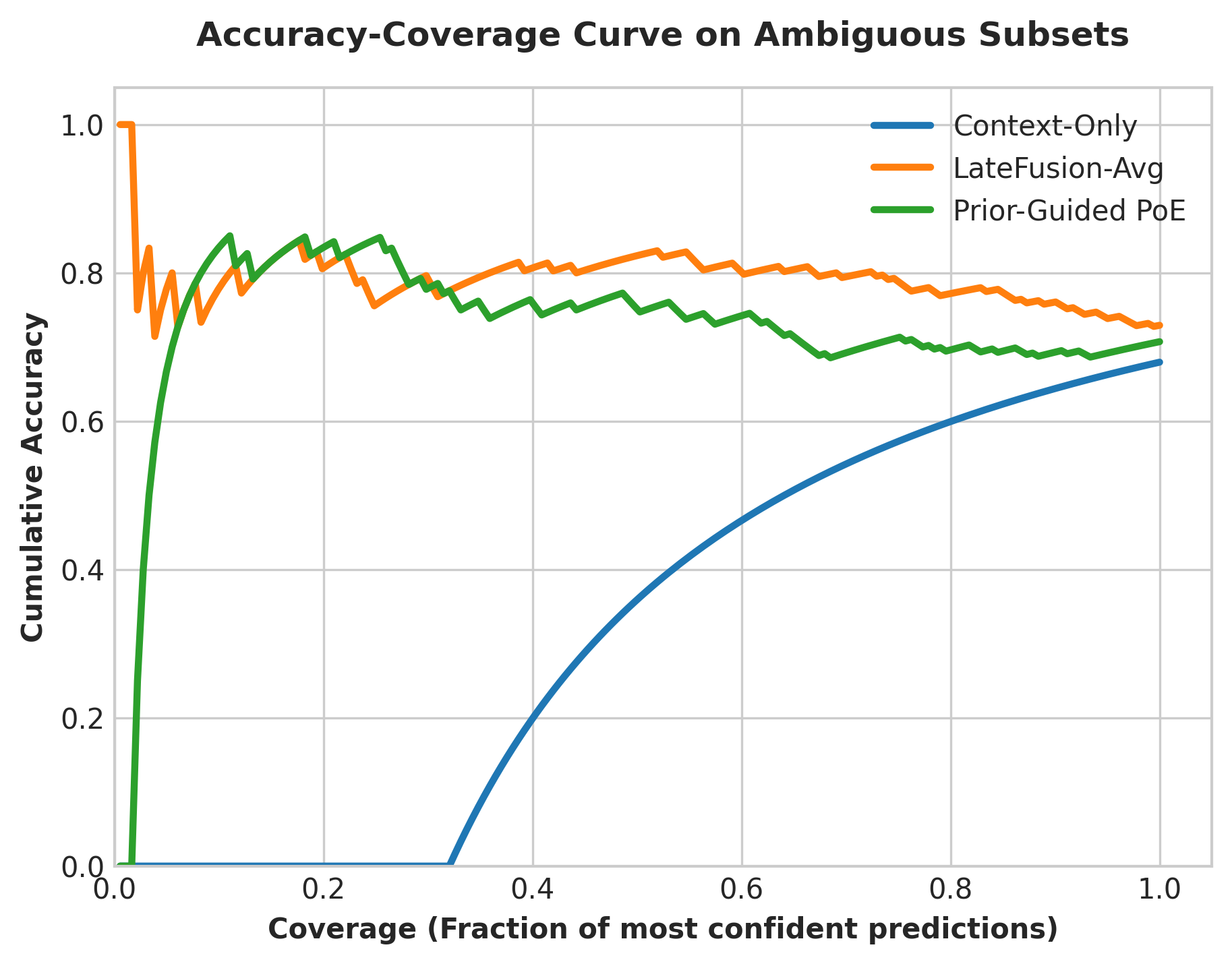}
    \caption{Accuracy--coverage curve on ambiguous subsets under tuned experts. Stronger fusion strategies substantially outperform context-only inference under ambiguity.}
    \label{fig:acc_cov}
\end{figure}

Figure~\ref{fig:acc_cov} shows the accuracy--coverage trade-off on ambiguous subsets (\texttt{near\_bowl} and \texttt{near\_door}). Compared with context-only inference, both stronger fusion strategies yield substantially better confidence-quality trade-offs under ambiguity. Late fusion attains the strongest cumulative accuracy over coverage, while the proposed PoE remains competitive and complements this result by achieving the best overall accuracy in the full evaluation.

% ------------------------------------------------
\section{Discussion and Limitations}

Our study is intentionally scoped as a proof-of-concept in a household feline setting. 
% As a result, the current experiments do not establish broad generalization across animals, homes, or species. Instead, they test a narrower claim: 
In a context-rich and behaviorally ambiguous household environment, treating context as prior knowledge can yield more reliable multimodal intent disambiguation than naive feature fusion.

The notion of intent in this work should be interpreted as an outcome-verified latent goal state associated with the observed clip.
% rather than as a fully verified internal mental representation. 
We also find that context and audio fusion are already strong, while the full model provides an additional gain. We hypothesize that this reflects the current home-collected dataset: vocalization cues are often highly discriminative, whereas pose-derived cues are limited by overlap in motion patterns and the stability of markerless keypoint extraction under large or blurred motion. Because clips without sufficiently valid keypoints are filtered out during preprocessing, some informative motion-heavy segments may be removed before feature extraction.

Finally, the current label space is intentionally small. Extending this formulation to richer intent sets, multi-home datasets, and cross-species settings is an important direction for future work.

% ------------------------------------------------
\section{Conclusion}

We presented CatSignal, a Bayesian-inspired context-gated Product-of-Experts framework for intent inference under strong contextual priors. Rather than treating spatial context as just another feature, our approach models context as a prior-like constraint, poses, and audio observations as complementary evidence, yielding a posterior-like distribution over latent intent. In a household cat testbed, this formulation improves overall accuracy, reduces context-driven shortcut failures, and remains competitive in uncertainty-sensitive analysis under ambiguity. More broadly, our results suggest that behavior understanding for non-speaking agents may be better approached as probabilistic intent inference than as conventional multimodal classification in settings where context sharply constrains feasible goals but does not fully resolve behavioral ambiguity.

\end{document}